\documentclass{IEEEtran4PSCC}
\usepackage[hidelinks]{hyperref}
\ifCLASSINFOpdf
  \usepackage[pdftex]{graphicx}
\else
\fi
\usepackage[cmex10]{amsmath}
\usepackage{xcolor}

\usepackage{multirow}

\usepackage{dsfont}
\usepackage{amssymb}
\newcommand{\STAB}[1]{\begin{tabular}{@{}c@{}}#1\end{tabular}}

\makeatletter
\let\old@ps@headings\ps@headings
\let\old@ps@IEEEtitlepagestyle\ps@IEEEtitlepagestyle
\def\psccfooter#1{%
    \def\ps@headings{%
        \old@ps@headings%
        \def\@oddfoot{\strut\hfill#1\hfill\strut}%
        \def\@evenfoot{\strut\hfill#1\hfill\strut}%
    }%
    \def\ps@IEEEtitlepagestyle{%
        \old@ps@IEEEtitlepagestyle%
        \def\@oddfoot{\strut\hfill#1\hfill\strut}%
        \def\@evenfoot{\strut\hfill#1\hfill\strut}%
    }%
    \ps@headings%
}
\makeatother

\begin{document}
%
\title{Hierarchical Demand Forecasting Benchmark for the Distribution Grid}

\author{
\IEEEauthorblockN{Lorenzo Nespoli, Vasco Medici}
\IEEEauthorblockA{ISAAC - SUPSI\\
Lugano, Switzerland\\
\{lorenzo.nespoli, vasco.medici\}@supsi.ch}
\and
\IEEEauthorblockN{Kristijan Lopatichki}
\IEEEauthorblockA{DESL - EPFL\\
Lausanne, Switzerland}
\and
\IEEEauthorblockN{Fabrizio Sossan}\\
\IEEEauthorblockA{PERSEE - Mines ParisTech\\
Sophia-Antipolis, France\\}
}


\maketitle

\begin{abstract}
We present a comparative study of different probabilistic forecasting techniques on the task of predicting the electrical load of secondary substations and cabinets located in a low voltage distribution grid, as well as their aggregated power profile. The methods are evaluated using standard KPIs for deterministic and probabilistic forecasts. We also compare the ability of different hierarchical techniques in improving the bottom level forecasters' performances. Both the raw and cleaned datasets, including meteorological data, are made publicly available to provide a standard benchmark for evaluating forecasting algorithms for demand-side management applications.
\end{abstract}

\begin{IEEEkeywords}
Forecasting, benchmark, hierarchical forecasting, electric demand.
\end{IEEEkeywords}

\thanksto{This project is carried out within the frame of the Swiss Centre for Competence in Energy Research on the Future Swiss Electrical Infrastructure (SCCER-FURIES) with the financial support of the Swiss Innovation Agency (Innosuisse - SCCER program) and of the Swiss Federal Office of Energy (project SI/501523).}

\section{Introduction}
The increasing monitoring capacity in low voltage (LV) and medium voltage (MV) distribution systems allows operators to gather power measurements from different levels of aggregation within the power grid. For instance, smart meters provide measurements from single households or buildings, dedicated power meters or phasor measurement units from secondary substations, and remote terminal units from primary substations at the interface between distribution and (sub)transmission systems. Real measurements of the power-flows “naturally” embed the notion of hierarchy. E.g., in a radial distribution system, the power flow at the grid connection point is, at the net of grid losses, the sum of the downstream elements. In the case of forecasts, however, the forecasted top-level series computed by using the information at that level of aggregation does not necessarily correspond to the sum of the bottom-level forecasts, thus invalidating the principle of hierarchy.  The process of re-establishing coherency between upper and aggregated bottom-level predictions is called reconciliation.

In current power systems operational practices, forecasts of the demand for a given aggregation level are generally computed by using measurements from that same level. Computing a top-level forecast by aggregating series at the bottom level is generally not pursued because bottom-level measurements are affected by higher levels of volatility, that impact negatively on forecasting performance. Moreover, the separation of concerns between different grid operators and data ownership conflicts do not encourage the exchange of data and the use of reconciliation strategies. 

However, future operational paradigms in active distribution networks will require tighter coupling between operations at different aggregation levels. The operator of an active distribution network will control distributed energy resources (i.e., demand response, storage, distributed renewable generation) to respect operational and physical constraints of the local power network (i.e., assure adequate voltage levels and respect line ampacity constraints) as well as providing ancillary services to the upper-level grid (i.e., dispatch, reserve, frequency control) through aggregation. In this context, the operator can take advantage of both disaggregated measurements and measurements at the grid connection point to compute coherent forecasts which satisfy the principle of aggregation to feed into optimal scheduling algorithms for the flexible resources.

In this paper, we first perform a comparison between different forecasters of the electrical demand. Then, based on the best performing method, we assess the effect of reconciliation on the forecasting performance. The analyses are carried out using data from an urban distribution network in Switzerland. The adopted data also includes numerical weather predictions (NWP) from a commercial provider. Compared to existing analyses and data sets in the literature that consider measurements at a 1-hour time resolution, e.g. \cite{Hong2014a} and \cite{Hong2019}, we use a resolution of 10 minutes, that is more in-line with the targets for real-time market operations. The resolution of the dataset, the presence of NWP variables, the completeness of the data and the variety of measured signals (described in details in appendix \ref{app:dataset}) makes this dataset unique and well suited for a forecasting benchmark for distribution grids. For this reason, we make the data used for this research publicly available in a repository \cite{public_dataset}.

The structure of the paper is as follows. Section II introduces the problem statement, Section III describes the adopted forecasting models, Section IV describes the tested reconciliation strategies, Section V presents and discusses results, and Section VI draws the conclusions.

\section{Problem formulation and case study}
\subsection{Problem Statement}
To illustrate the problem, we consider the grid in Fig.~1. It is a radial system that interfaces five nodes to the grid connection point (GCP), which is connected to the upper-level grid through a transformer. Each node corresponds to a specific active power injection (e.g., demand or generation), which is measurable. In this paper, we assume that the grid losses are negligible, so the active power at the grid connection is the algebraic sum of the nodal injections. Based on the historical measurements, the operator can determine forecasts for all the nodes, including those at the GCP. While real power measurements will embed the hierarchical structure imposed by the grid topology (i.e., the power at the GCP will match the sum of individual nodes), forecasts will not as they are estimated individually. The problem that we tackle in this paper is how to forecasts the nodal injections individually, and secondly, how to reconcile them.

\begin{figure}[!ht]
\centering
\includegraphics[width=0.5\columnwidth]{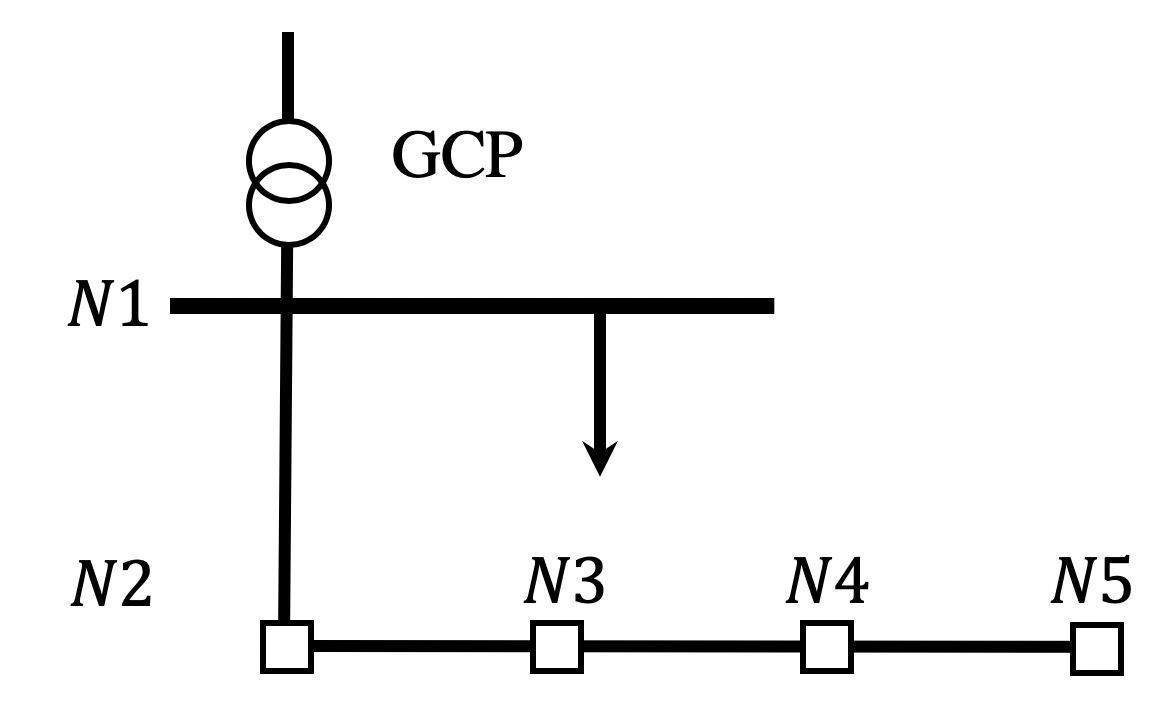}
\caption{Schmatic view of the monitored grid.}
\end{figure}

\subsection{Input data}
The input data consists of power measurements and meteorological forecasts relative to a set of power meters located in Rolle (Switzerland). The available measurements are thoroughly described in Appendix~1. In total, we consider 24 nodes, with an average power of 81 kW, and 7 synthetic aggregated series that are created by partial aggregations of the original data, as explained in section \ref{sec:hier}.
The complete dataset can be downloaded at \cite{public_dataset}.

\subsection{Forecast formulation}\label{sec:formulation}
We compare the accuracy of different parametric and non-parametric forecasting techniques using k-fold cross-validation (CV). As the power generation gets more decentralized and uncertain due to the presence of renewable energy, system operators have moved from day-ahead optimization (as for standard unit commitment problem) to shorter clearing times, solving the optimization problem in a receding horizon fashion \cite{Zheng2015a}.

For this reason, we tested the methods using the same sliding window concept. We applied a preliminary causal embedding of the explanatory variables and the target time series, which we explain in the following.
Starting from the original time series $s\in \mathcal{S}$, a predictors (or regressors) matrix $X$ and a target matrix $Y$ are obtained. Given a dataset with $T$ observations, a prediction horizon of $h$ steps ahead, and an history embedding of $e$ steps,  we obtain the Hankel matrix of targets $Y\in \rm I\!R^{ (T-h-e) \times h}$, and the Hankel matrix of the past regressors, $X_p\in \rm I\!R^{ (T-h-e) \times n_x e}$, where $n_x$ is the number of regressors. Verbosely, $X_p$ and $Y$ can be written as:
\begin{equation*}
X_p = \left[\begin{smallmatrix}
x_{1,t-e}&x_{1,t-e+1}&...&x_{1,t}&x_{2,t-e}&...&x_{n_x,t}\\
&&&...&&&\\
x_{1,t-e+1}&x_{1,t-e+2}&...&x_{1,t+1}&x_{2,t-e+1}&...&x_{n_x,t+1}\\
x_{1,T-2h}&x_{1,T-2h+1}&...&x_{1,T-h}&x_{2,T-2h}&...&x_{n_x,T-h}\\ 
\end{smallmatrix}\right]
\end{equation*}
\begin{equation*}
Y = \left[\begin{smallmatrix}
y_{t+1}&y_{t+2}&...&y_{1,t+h}\\
&&&...&&&\\
y_{T-h+1}&y_{T-h+2}&...&y_{T}\\
\end{smallmatrix}\right]
\end{equation*}
where $x_{1,t}$ stands for the first regressor at time $t$.
In hour case, we fixed $h=144$, corresponding to a prediction horizon of 24 hours ahead. The past regressors matrix $X_p$ is then augmented with categorical time features, e.g. day of week, and NWP variables, to obtain the final regressors matrix $X$. 
Rows of the $X$ and $Y$ matrices are then used to create the cross-validation training and testing dataset folds, $\{\left(D_{tr,f},D_{te,f}\right), f = 1,2,\dots,k \}$, where $k$ is the number of folds. Since we are dealing with time series forecasting, in order to avoid having very similar entries in some of the $D_{tr,f}$ and $D_{te,f}$ rows, we built them such that they are always separated by the embeddig length. The procedure we have used to build the different folds is the following. Each fold is divided into 10 days sequences, whose first 7 belong to the training set $D_{tr,f}$. Since for most of the regressors we have adopted a 24 hours embedding, for each sequence in $D_{tr,f}$ we discarded the 8-th and 10-th day, while the 9-th day is assigned to the testing dataset $D_{te,f}$. 

\begin{figure}[!ht]
\centering
\includegraphics[width=2.5in]{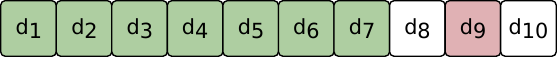}
\caption{Cross validation segment. Green squares: training days. Red square: test day. During testing, due to the adopted 24 hours embedding, the algorithms only see data contained in the 8-th day, avoiding overlapping of training and testing datasets.}
\label{fig:cv}
\end{figure}

An example of the division of a data sequence in training and testing days is shown in Fig. \ref{fig:cv}. We then adopted a 10 fold CV ($k=10$), for each of which we shifted the start of the sequences by one day. In this way, by stacking the prediction of all the folds, it is possible to obtain forecasts for the whole period of the original dataset.

\section{Forecasting models}\label{sec:methods}
Since the scope of the paper is to provide a benchmark, we didn't introduce any new forecasting algorithms. On the contrary, we focused on those methods having a long track records in providing good results in the forecasting literature, and adapted them to be tested in cross validation, if needed.  

\subsection{ARMAX}
ARMAX are state-full models, i.e. they require past values of both target variables and prediction error to perform a prediction for the next timestep, and have been largely applied to time series forecasting. The model is a regression where the covariates are a set of exogenous inputs $x \in \mathds{R}^{n_x}$, some past values of the target variable $y$, and the model error is assumed to be a white noise process $\epsilon$. The model can be written as:
\begin{equation}
    \phi(q) y = \beta x + \theta(q) \epsilon
\end{equation}
where $q$ is the backshift operator, $\phi(q) = 1 - \sum_{i=1}^{n_p} \phi_i q $, $\theta(q) = 1 - \sum_{i=1}^{n_q} \theta_i q $  and $n_p$ and $n_q$ are the auto regressive and moving average orders of the model.
Due to the stateful nature of the models, it is not possible to use all the observations in the CV data folds $D_{tr,f}$, since those are discontinuous. To overcome this issue, for each segment of 10 days (see Fig. \ref{fig:cv}) of a given training data fold, we fit a different ARMAX model. The fitted models are then used to form an ensemble, i.e. the final prediction is given by:
\begin{equation}
    \hat{y}_t = \frac{1}{n_s} \sum_{i=1}^{n_s} \hat{y}_{t,i}    
\end{equation}
where $\hat{y}_{t,i}$ is the prediction of the $i$-th model and $n_s$ is the number of segments in the current CV data fold $D_{tr,f}$. The ensemble process invalidates the assumptions upon which the confidence interval of the predictions are usually derived. For this reason, we have obtained the prediction quantiles a-priori, for given combination of time of the day and step ahead. Denoting as $q_{\alpha_i}$ as the empirical $\alpha$-quantile: 
\begin{equation}\label{eq:emp_quant}
\hat{q}_{\alpha_i,h,d} =  q_{\alpha_i}(e_{h,d}) 
\end{equation}
where $e_{h,d}$ is the set of training errors obtained on $D_{tr,f}$ related to the $h$-th step ahead and to the $d$-th step of the day.
The optimal values of the autoregressive and moving average orders, $n_p$ and $n_q$, are obtained using the \texttt{autoarima} R package using random samples from the time series. The resulting values, respectively 6 and 5, were kept fixed during the CV. Note that the actual models' parameters have still been properly fitted in CV; in our case, the ARMAX's orders represent hyper-parameters of the overall model, and this procedure can be seen as fixing them to an `educated guess'. 

\subsection{Detrended Holt Winters}
The Holt-Winters (HW) model \cite{Holt2004} is a special class of the exponential smoothing \cite{Gardner1985}, which consists of three smoothing equations, such that the final prediction is a combination of the level $a$, trend $b$ and seasonality $s$. We tested different flavors of the HW families and based on performance, we adopted a double seasonality additive HW:
\begin{equation}\label{eq:HW}
\begin{aligned}
\hat{y}_{t+h} &= (a_t +hb_t) +s_{1,t-p(1)+1+(h-1)\backslash p_1}+s_{2,t-p_2+1+(h-1)\backslash p_2}\\
a_t &= \alpha(y_t-s_{1,t-p_1}-s_{2,t-p_2}) + (1-\alpha)(a_{t-1} + b_{t-1})\\
b_t &= \beta(a_t - a_{t-1}) + (1-\beta)b_{t-1}\\
s_{1,t} &= \gamma_1(y_t-a_t-s(2,t-p_2))  + (1-\gamma_1)s_{1,t-p_1}\\
s_{2,t} &= \gamma_2(y_t-a_t-s(1,t-p_1))  + (1-\gamma_1)s_{2,t-p_2}\\
\end{aligned}
\end{equation} 
where $\alpha$, $\beta$, $\gamma_1$ and $\gamma_2$ are parameters to be learned from data, while $p_1=96$ and $p_2=672$ are the periods of the seasonalities, and $\backslash$ is the modulo operator. The values for $p_1$ and $p_2$ correspond to a daily and weekly period. The model \eqref{eq:HW}, and HW in general, do not include exogenous inputs. Since quantities like external temperature and irradiance are important explanatory variables in load forecasting, we included them with an a-priori linear detrend, such that the new target is $y = y - X\beta_d $, where $X$ is a three column matrix containing $GHI$, $T$ and the unit vector (for the intercept), and $\beta_d$ is the vector of linear coefficients.
Usually, a single set of $\alpha$, $\beta$, $\gamma_1$ and $\gamma_2$ values is fitted, and the prediction of each step ahead is obtained applying equations \ref{eq:HW} recursively, as usually done for state-space systems. To increase the accuracy of the method, we instead fitted 144 sets of $\alpha$, $\beta$ and $\gamma$ parameters, based on the step ahead. As done for the ARMAX models, we used random samples from the bottom time series to fit these parameters. Due to the linear detrend we applied to the target, the fitted $\beta$ values were close to $0$ for all the steps ahead, and thus we decided to fix this parameter to $0$. Also, in this case, the prediction quantiles are obtained a priori, using equation \ref{eq:emp_quant}.

\subsection{K-nearest neighbors}
The K-nearest neighbours \cite{Hastie2009} regressor is based on a simple but effective technique. The method selects the most similar K points in the training set, based on the features at the given prediction time. A weighted average of the target value of the selected points is then used to obtain the final prediction. 
\begin{equation}
    \hat{y}_t = \sum_{i=1}^k \omega_i y_i
\end{equation}
where $\omega_i$ and $y_i$ are the weight and the target variable of the $i$-th neighbour, respectively. In our case, we have used the Euclidean distance as a similarity measure to select the neighbours, and the inverse distance as weights, as implemented in the \texttt{KNeighborsRegressor} class of \texttt{scikit-learn} Python package. Forecast quantiles are obtained estimating them from the distribution of the k nearest neighbours predictions. We adopted a multiple-input single-output (MISO) strategy, in which different models are trained for different steps ahead, for a total of 144 models for each fold.  

\subsection{Gradient Boosting}
Tree boosting is a widely used machine learning technique, both for classification and regression tasks. The method relies on repeatedly fitting regression trees on the residual of the predicted variable. In order to reduce overfitting, the well-known implementation of \texttt{XGBoost} \cite{Chen2016} includes a penalization on the number of parameters in the fitting process. In this comparison, we relied on the LightGBM implementation described in \cite{Ke2017}, which is characterized by a highly parallelizable algorithm for the construction of the trees, tailored to big datasets. Also in this case, we adopted a MISO strategy.

\subsection{Hierarchical forecasting} \label{sec:hier}
Hierarchical forecasting aims to increase the accuracy of the prediction of signals organized in a hierarchical structure with increasing levels of aggregations, with respect to the case in which the aggregated signals are forecasted directly. Secondly, it aims at providing aggregate-consistent forecats, which can be obtained by encoding the hierarchical structure in a learning algorithm. This is done exploiting the forecasts of the bottom series $y_b \in \mathds{R}^{T \times n_b}$: usually an optimization technique is used to find a latent variable $\tilde{y}_b \in \mathds{R}^{T \times n_b}$, which can be used to approximate the whole set of original forecasts $y = \left[y_u^T ,y_b^T \right]^T \in \mathds{R}^{T \times n}$, where $n = n_b + n_u$ and $n_b$ and $n_u$ are the number of the bottom and upper time series. Formally, the following must hold for $\tilde{y}_b$:
\begin{equation}
    \tilde{y}^T = S \tilde{y}_b^T
\end{equation}
where $S \in \mathds{R}^{n\times n_b}$ is a summation matrix and $\tilde{y}$ is the set of corrected forecasts.
In this paper we have fictitiously aggregated the bottom time series in order to provide two levels of aggregation, such that $S$ is:
\begin{equation}
S = \left[
\begin{matrix}
&\mathds{1}_{n_b}\\
&I_{2} \otimes \mathds{1}_{n_b/2}\\
&I_{4} \otimes \mathds{1}_{n_b/4}\\
&I_{n_b}
\end{matrix}
\right] 
\end{equation}
where $I_{k}$ is the identity matrix of dimension $k\times k$, $\mathds{1}_{k}$ is the unit raw vector of dimension $k$ and $\otimes$ is the Kronecker product.
Following this approach, in \cite{Hyndman2011} the authors used ordinary least squares (OLS) regression to reconcile the forecasts in the hierarchy. Elaborating on this approach, \cite{Wickramasuriya2017a,Wickramasuriya2018} proposed a trace minimization method (called minT) in which the covariance matrix of the forecasters error is estimated to perform a weighted least squares regression. In \cite{Taieb2017b}, an elastic net penalization was proposed in order to induce sparseness in the forecasters adjustments, and the benefit was shown on the reconciliation of the forecasts for the power consumption of residential consumers. A probabilistic hierarchical reconciliation through empirical copulas is proposed in \cite{Taieb2017a}. Another probabilistic reconciliation approach has been recently proposed in \cite{Corani}: under the hypothesis of a joint Gaussian distribution for the base forecasts, this method exploits Bayes rule to obtain a closed-form solution to the probabilistic reconciliation.

\section{Numerical results}
\subsection{Evaluation KPIs}
The results have been compared by means of standard key performance indicators (KPIs) for regression tasks. For the point forecast evaluation we have used the root mean squared error (RMSE) and the mean absolute percentage error (MAPE). The two aforementioned metrics have been evaluated using two levels of aggregation: we retrieved the expected value over the cross validation, as a function of the step ahead and hour of the day; secondly we have further aggregated the KPIs with respect to the hour of the day. Formally, we have evaluated:

\begin{align}
\mathrm{RMSE}_{d,h} &= \frac{1}{n_f}\sum_{f=1}^{n_f}  \left( \frac{1}{\vert \mathcal{J}_{h,d,f} \vert} \sum_{j\in\mathcal{J}_{h,d,f}} e_{j}^2 \right)^{1/2}\label{eq:RMSE_map}\\
\mathrm{MAPE}_{d,h} &= \frac{100}{n_f}\sum_{f=1}^{n_f}  \left( \frac{1}{\vert \mathcal{J}_{h,d,f} \vert} \sum_{j\in\mathcal{J}_{h,d,f}} \frac{abs(e_{j})}{y_{j}} \right) \label{eq:MAPE_map}
\end{align}

where $\mathcal{J}_{h,d,f}$ is the set of observations relative to the $h$-th step ahead, $d$-th step of the day and $f$-th CV fold, $e$ is the forecast error, $n_f$ is the number of folds.

The resulting $\mathds{R}^{n_d, h}$ matrices are then normalized with the values of the same KPIs obtained using the persistence model. The probabilistic forecasts have been evaluated by means of the time average of the quantile loss $\bar{l}_{\alpha}$, and the quantile score $QS(\hat{q}_{\alpha},y)$, which is a proper scoring rule \cite{Gneiting2007a,Bentzien2014}, and it's defined as the expected quantile loss: 
\begin{align}
\epsilon_{\alpha} &= \hat{q}_{\alpha} - y \\ 
\bar{l}_{\alpha} &= \sum_{t=1}^T \epsilon_{\alpha} \left(\mathds{I}_{\epsilon_{\alpha}\geq 0} -\alpha \right)\\
QS &= \int_0^1 \bar{l}_{\alpha} \mathrm{d} \alpha 
\end{align}

where $\hat{q}_{\alpha}$ is the predicted $\alpha$-quantile, while $y$ is the observed ground truth.

\subsection{Single time series forecasting}
\begin{figure}[!ht]
\centering
\includegraphics[width=3.5in]{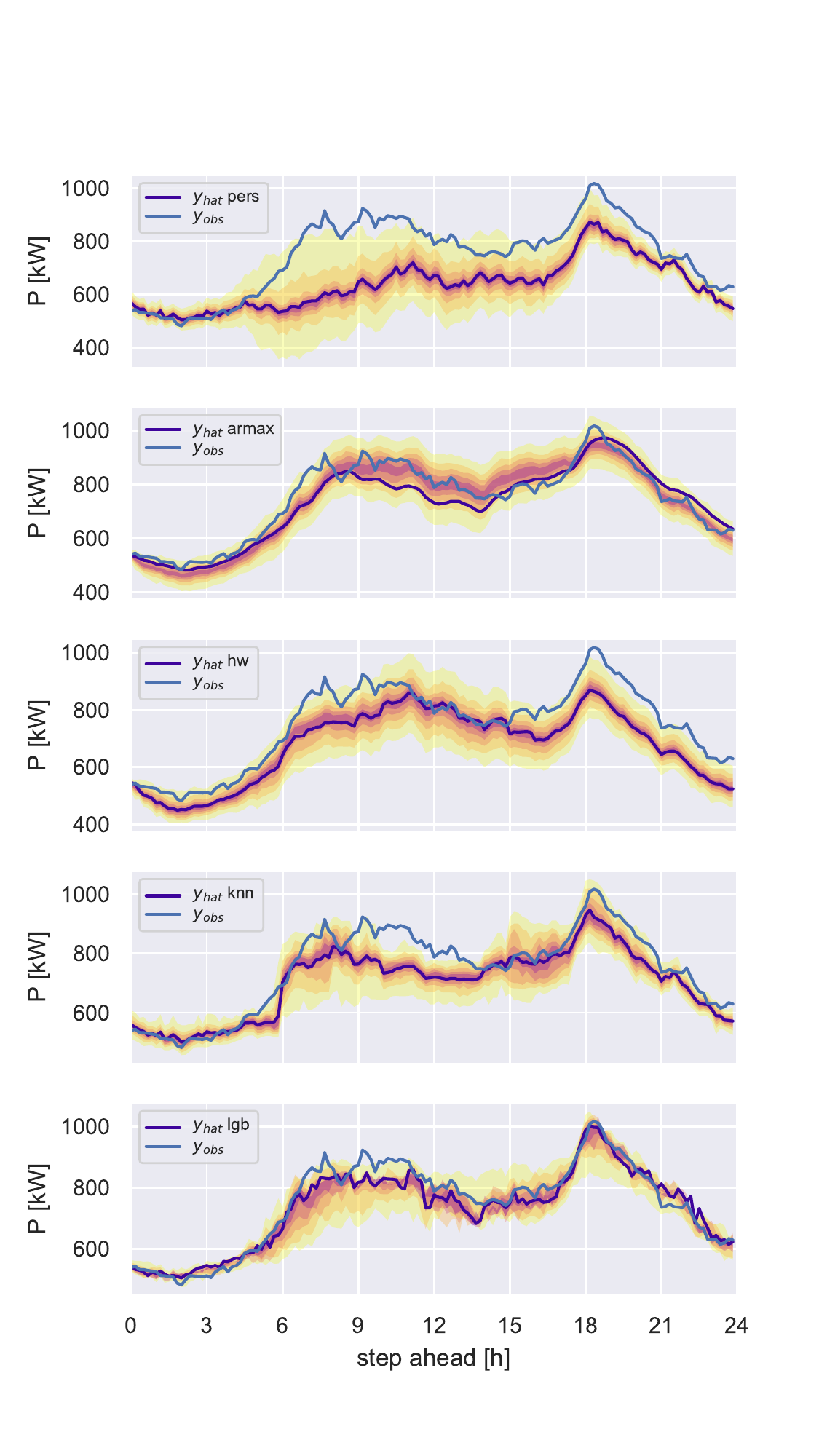}
\caption{Example of day-ahead forecasts for the whole power aggregate, and for the different forecasters methods.}
\label{fig:example_ts}
\end{figure}

We performed day-ahead forecasts for all the time series in the hierarchy previously described, applying the methods introduced in section \ref{sec:methods} and following the CV approach introduced in section \ref{sec:formulation}. An example of day-ahead forecasts for the whole aggregate is shown in Fig. \ref{fig:example_ts}, along with eleven evenly spaced quantiles in the $\left[0.05,0.95\right]$ interval. From the picture, it can be noticed that the prediction of the ARMAX model is not centred in its quantile prediction during the central part of the day. This means that, during these hours, the model consistently underestimated the load in the training sets, and the empirical estimation of quantiles using equation \eqref{eq:emp_quant} reports this effect, which couldn't be visible using the standard Gaussian process assumption.   

\begin{figure}[!ht]
\centering
\includegraphics[width=3.5in]{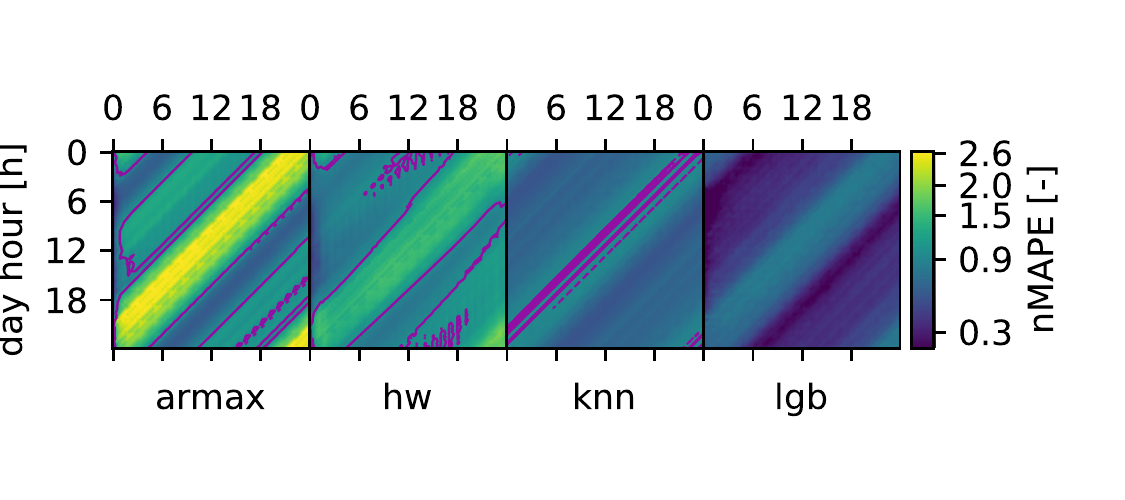}
\caption{Average MAPE from CV, normalized with the persistence forecaster MAPE, plotted as a function of day hour (vertical axis) and hour ahead time of the prediction (horizontal axis). The regions inside violet contours are the ones in which the persistence model has a better MAPE w.r.t. the considered forecaster (nMAPE$\geq1$) .}
\label{fig:MAPE_maps}
\end{figure}

\begin{figure}[!ht]
\centering
\includegraphics[width=3.5in]{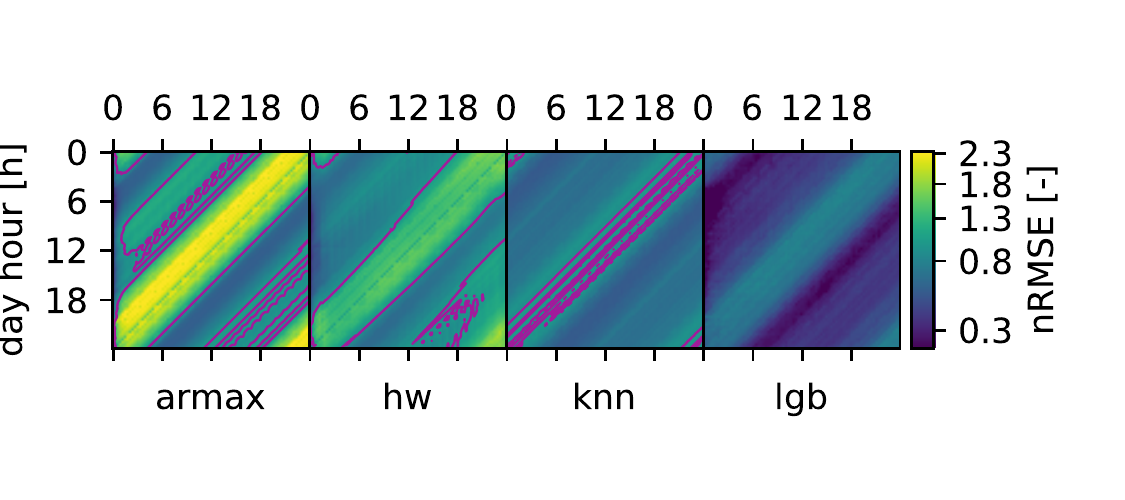}
\caption{The same kind of plot of figure $\ref{fig:MAPE_maps}$ is shown, with RMSE values.}
\label{fig:RMSE_maps}
\end{figure}

In Fig. \ref{fig:MAPE_maps} and \ref{fig:RMSE_maps} the normalized RMSE and MAPE matrices of equation \eqref{eq:RMSE_map} and \eqref{eq:MAPE_map} are reported for the tested forecasters. The regions for which the values exceed the unity, that is, where the persistence method achieves better performances, are enclosed in a violet line. For all the methods, we can see that the combination of step ahead and step of the day close to the antidiagonal present the highest values of normalized KPIs. This means that in a time window of a few hours centred around midnight, the persistence method is already very accurate. The KNN and LightGBM models are strictly better than the persistence model for all the steps ahead and for all the times of prediction.

\begin{figure}[!ht]
\centering
\includegraphics[width=3.5in]{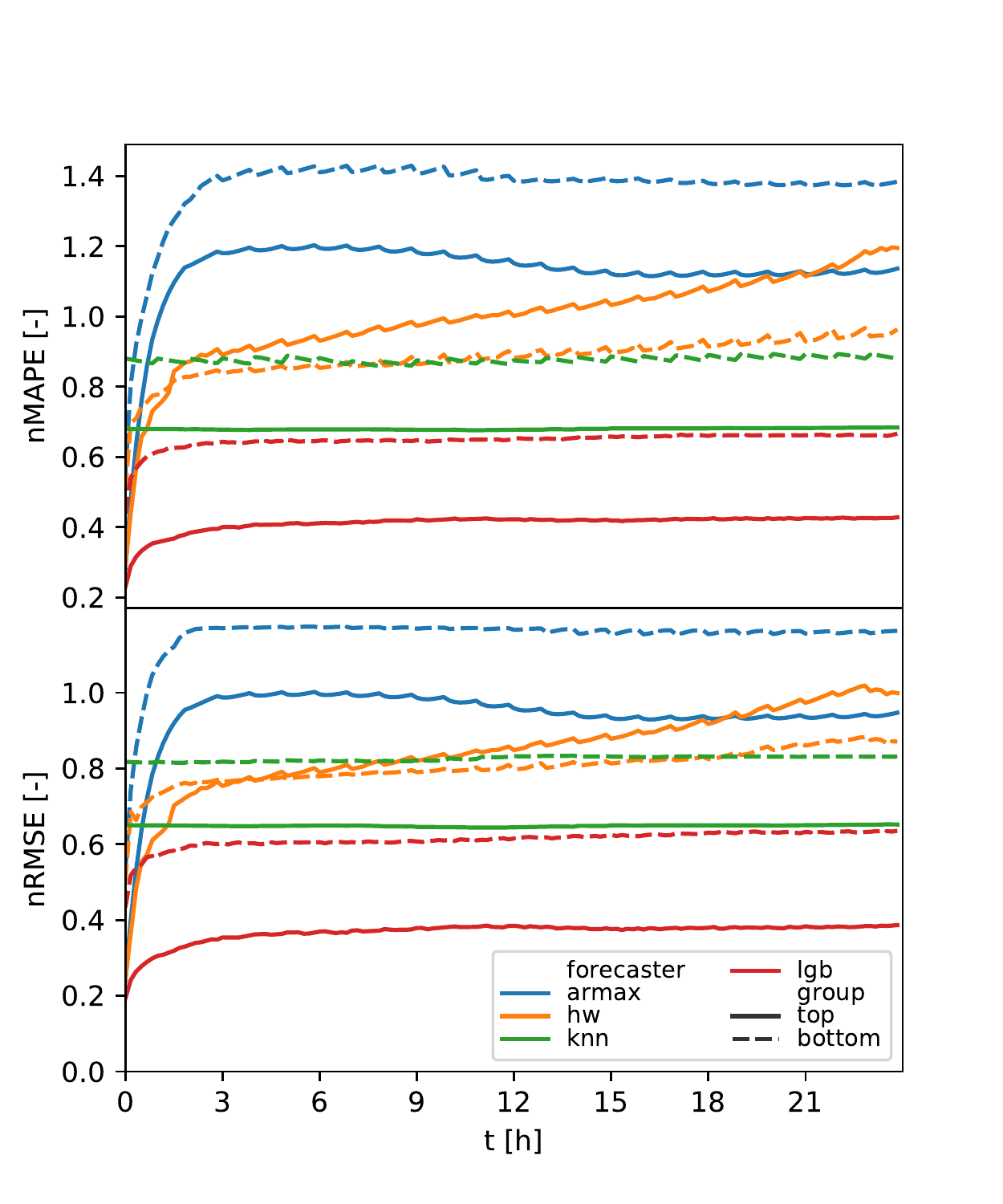}
\caption{nMAPE and nRMSE of different forecasters. Continuous lines: values refer to the top series (whole aggregate). Dashed lines: values refer to the average on the bottom time series.}
\label{fig:nMAPEnRMSE}
\end{figure}

Fig. \ref{fig:nMAPEnRMSE} shows the raw average of the normalized RMSE and MAPE matrices of equations \eqref{eq:RMSE_map} and \eqref{eq:MAPE_map}, i.e. the sample expectations of these KPIs with respect to the prediction horizon. The dashed lines represent the average over all the bottom series, while the continuous lines refer to the whole aggregate. It can be seen how, while all the methods are strictly better than the persistence model in the first few step ahead, the ARMAX model rapidly worsen its performance, especially when considering the bottom series. On the contrary, the HW model shows better performances on the bottom series, while being strictly better than the persistence model in terms of RMSE for all the steps ahead. The KNN and LightGBM models are consistently better for all the prediction horizons for both the bottom and top series. However, we can see how the KNN method is not able to obtain low scores for the first prediction steps. This is because the KNN model does not include dynamics and is not able to discriminate the importance of the covariates based on the prediction step, while this is the case for LightGBM, being a tree-based model.   

\begin{figure}[!ht]
\centering
\includegraphics[width=3.5in]{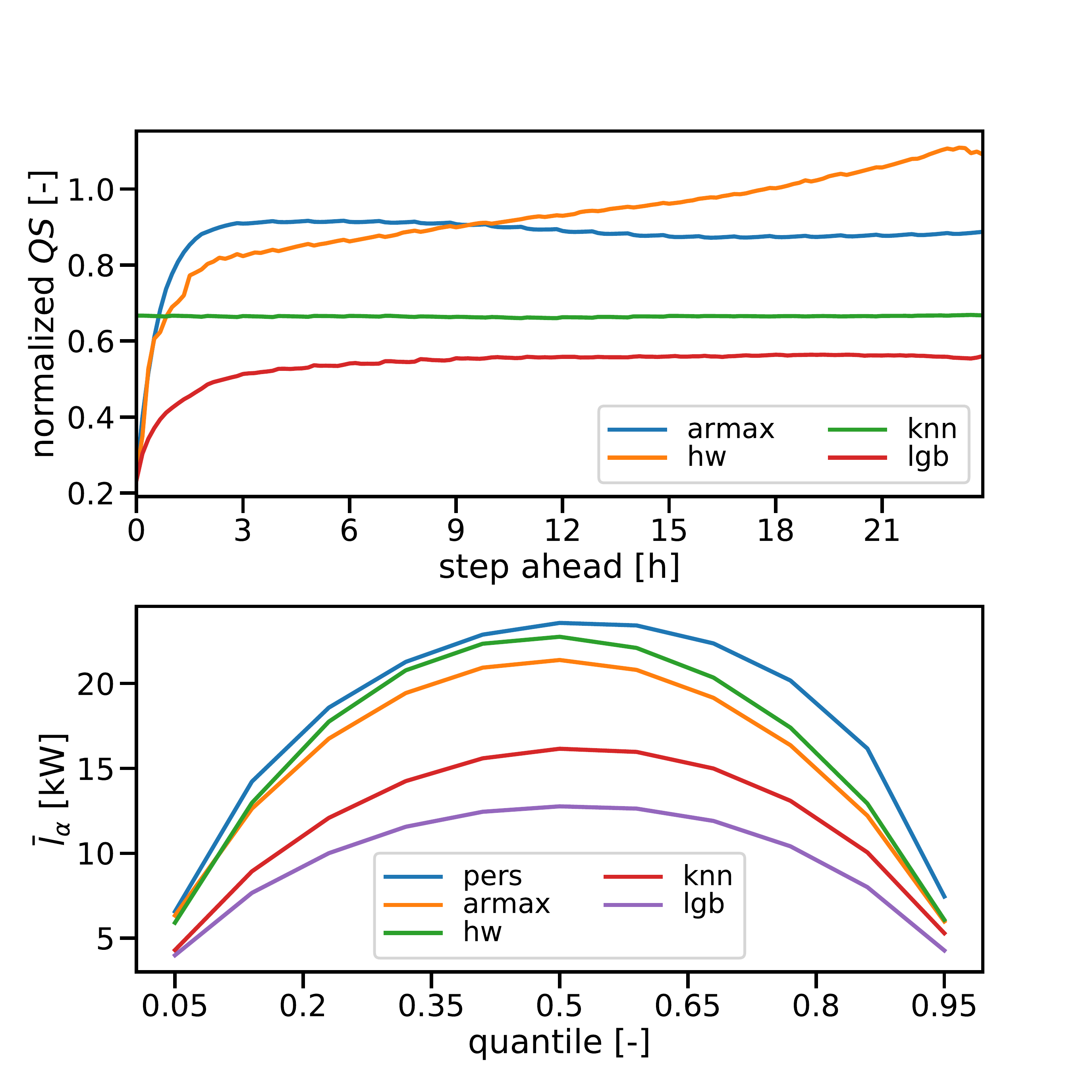}
\caption{Top: average normalized $QS$ as a function of step ahead for different forecasters. Bottom: mean quantile loss as a function of the quantile for different forecasters.}
\label{fig:quantile_scores}
\end{figure}

In Fig. \ref{fig:quantile_scores} both the normalized $QS$ as a function of step ahead and the mean quantile loss as a function of the predicted quantile, are shown for the whole aggregate. The upper part of the figure shows that the HW method presents less reliable predicted quantiles after 10 hours ahead, while the other methods present a lower $QS$ with respect to the persistence model, for all the prediction horizons. The lower part of the figure shows $\bar{l}_{\alpha}$ mediated on all the prediction horizon. In this case, all the methods achieve better results compared to the persistence model. Once again, the ranking of the forecasters is unchanged, with the non-parametric models achieving better results. 
Table \ref{table2} summarizes the average scores for the different forecasters, reporting the time-averaged MAPE and RMSE. The number on the left refers to the aggregated power profile, while the one on the right refers to the average score over the bottom time series. We can see that the best scores are always obtained using the LightGBM model. 

\begin{table}[!ht]
    \renewcommand{\arraystretch}{1.3}
    \centering
    \caption{Mean scores for the aggregated time series (left) and for the bottom series (right)}
    \label{table2}
    \begin{tabular}{cc|c|c|c|c|l}
        \cline{3-6}
        & & \multicolumn{4}{ c| }{Forecaster} \\ \cline{3-6}
        & & armax & hw & knn & lgb  \\ \cline{1-6}
        \multicolumn{1}{ |c  }{\multirow{2}{*}{\STAB{\rotatebox[origin=c]{90}{Score $\quad$}} }} &
        \multicolumn{1}{ |c| }{\STAB{\rotatebox[origin=c]{90}{ MAPE }}} 
        &  8.2 / 21.8 & 7.2 / 14.9 & 4.9 / 13.8 & \textbf{3.0} / \textbf{9.8}
        \\ \cline{2-6}
        \multicolumn{1}{ |c  }{}                        &
        \multicolumn{1}{ |c| }{\STAB{\rotatebox[origin=c]{90}{ RMSE }}} 
        &  67.1 / 6.1  & 60.4 / 4.7 & 46.2 / 4.4 & \textbf{26.2} / \textbf{3.1}  \\ \cline{1-6}
    \end{tabular}
\end{table}

\subsection{Hierarchical reconciliation}
\begin{figure}[!ht]
\centering
\includegraphics[width=3.5in]{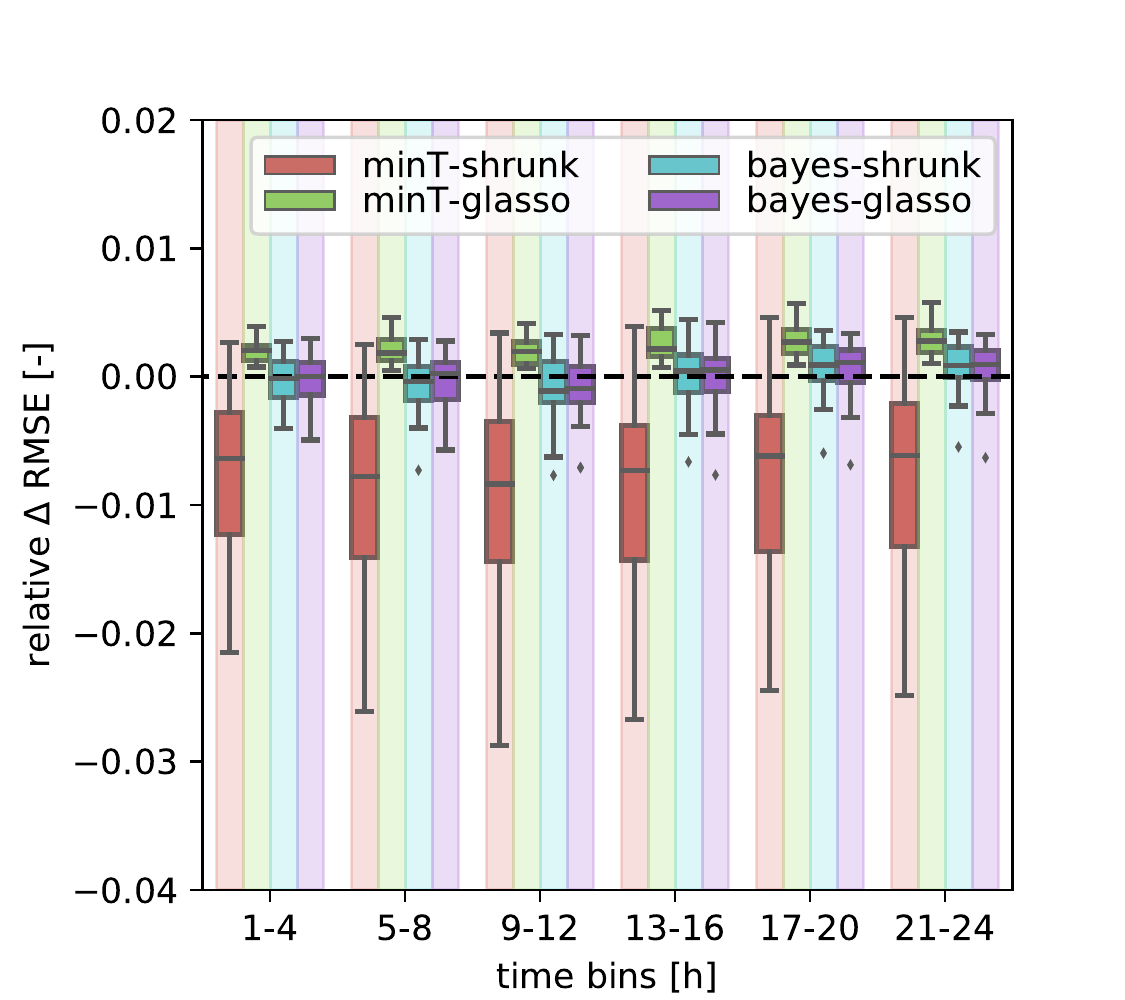}
\caption{Boxplots of the RMSE reduction for the bottom time series, using different reconciliation techniques, as a function of step ahead. The values are normalized with the RMSE of the base forecasters, and aggregated using 4 hours bins. Positive values indicate an improvement.}
\label{fig:HR_bottom_boxplots}
\end{figure}

\begin{figure}[!ht]
\centering
\includegraphics[width=3.5in]{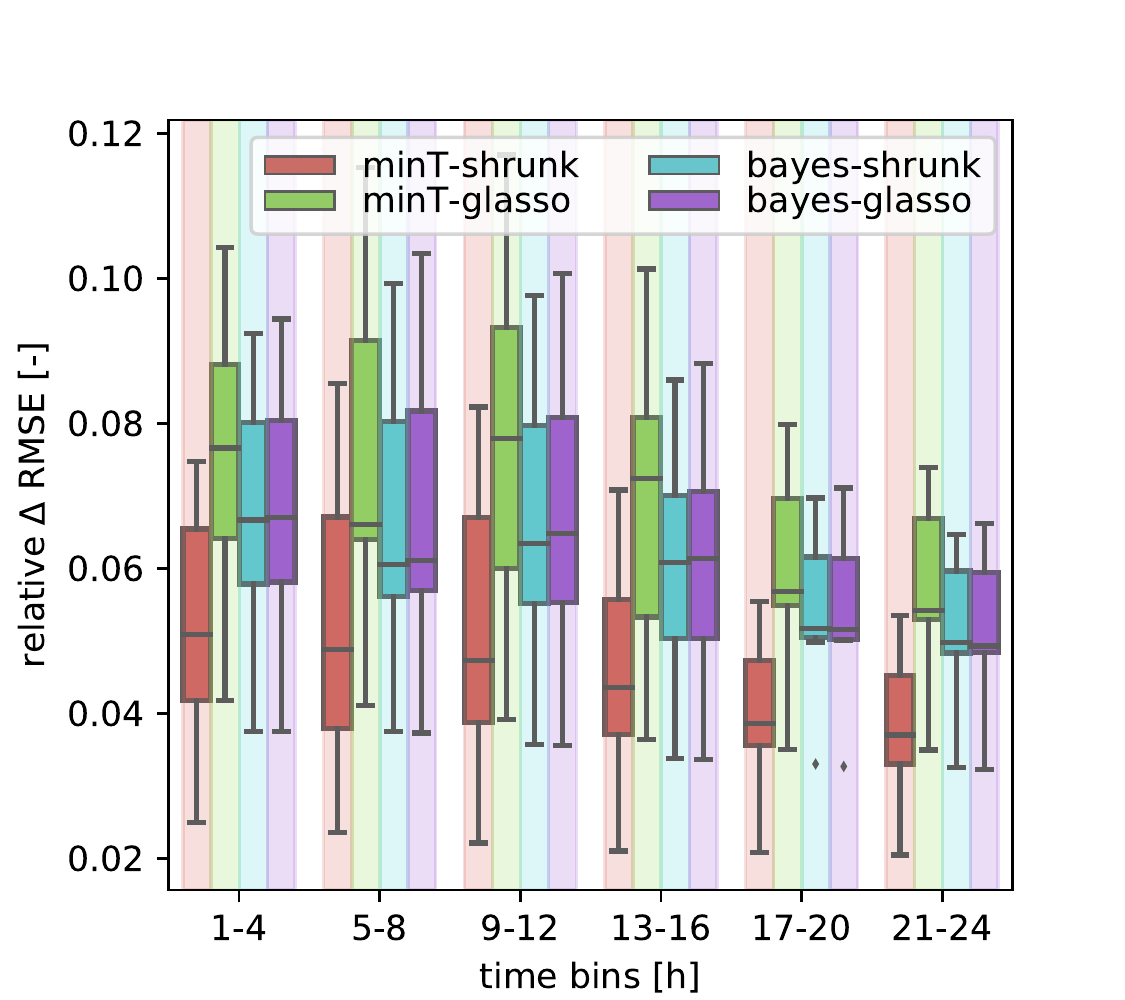}
\caption{The same kind of plot of figure \ref{fig:HR_bottom_boxplots}, referred to the aggregated time series.}
\label{fig:reconciliation_rRMSE}
\end{figure}

For this analysis, we use the LGBM forecaster, that was the best performing model on the single time series under all KPIs, as discussed above. We retrieve the base forecasts for the whole dataset for all the 24 bottom series and the additional 7 aggregations using the CV method explained in section \ref{sec:formulation}. As introduced in Section~\ref{sec:hier}, we test the minT \cite{Wickramasuriya2017a} and Bayesian \cite{Corani} methods in combination with two different techniques for the estimation of the error covariance matrix, on which both the methods rely to obtain the reconciled time series. We tested both the Ledoit-Wolf shrinkage approach\cite{Ledoit2004} and the graphical Lasso method \cite{Friedman2008} using the implementation in the \texttt{scikit-learn} Python package.  
Figures \ref{fig:HR_bottom_boxplots} and \ref{fig:reconciliation_rRMSE} show the relative reduction of RMSE compared to the base forecasts for the bottom and aggregated time series, respectively. The results are presented by means of temporal aggregations of 4 hours each, with respect to the step ahead. As it can be seen from Fig. \ref{fig:HR_bottom_boxplots}, on the one hand, the combination of minT with the shrunk covariance estimation score the worst performance and it even leads to increasing the RMSE on the bottom time series. On the other hand, minT with graphical Lasso covariance estimation provides the best results. The Bayesian reconciliation showed less sensitivity to the adopted covariance estimation method. However, the contribution of the reconciliation on the reduction of RMSE is marginal since it is lower than 1\% in all cases.

As Fig. \ref{fig:reconciliation_rRMSE} shows, the reconciliation displayed a higher reduction on the RMSE on the aggregated series forecasts. Also in this case, minT with shrunk covariance scores the worst, whereas minT with graphical Lasso is the best performing model.

\begin{figure}[!ht]
\centering
\includegraphics[width=3.5in]{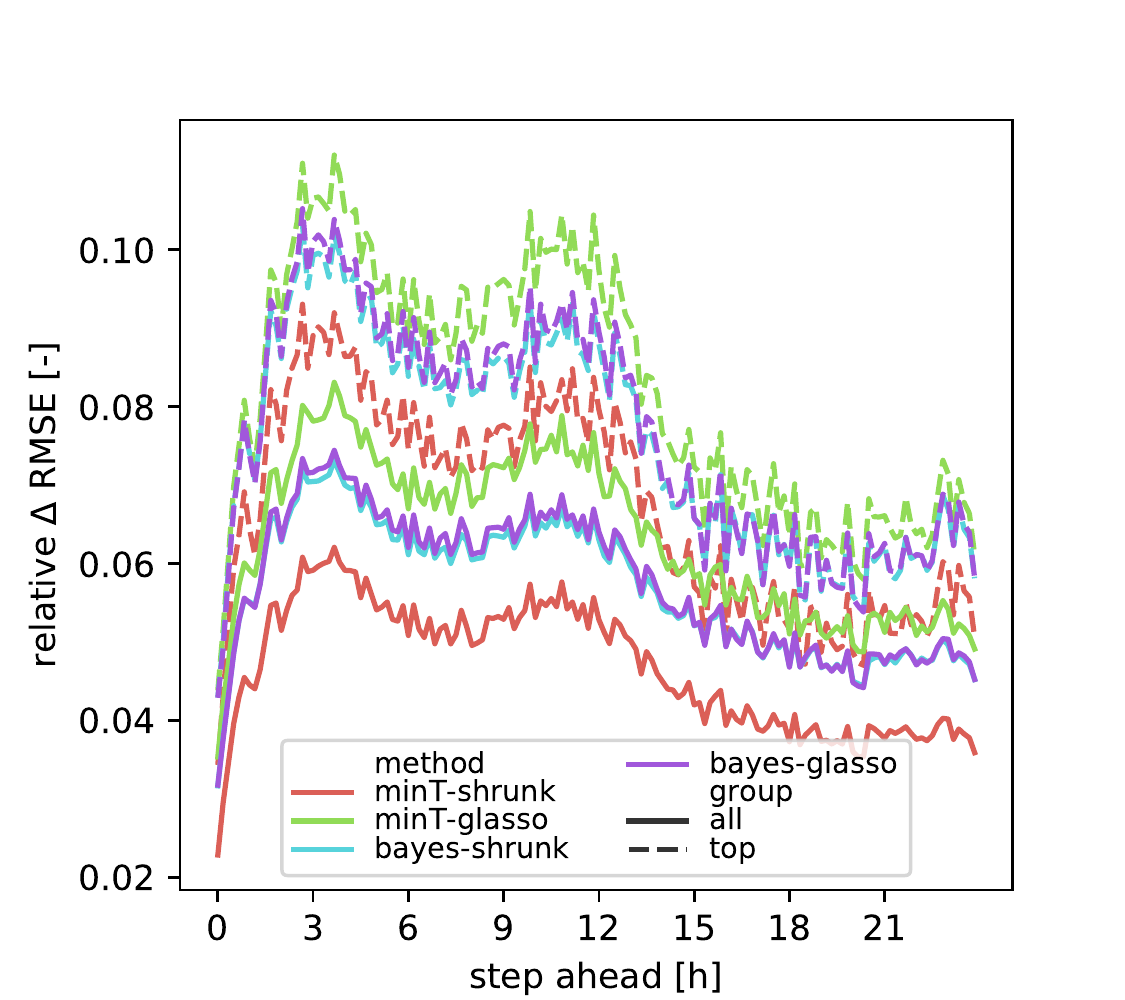}
\caption{Relative RMSE reduction as a function of step ahead. Continuous lines: average value over the whole hierarchy. Dashed lines: top series (whole aggregate).}
\label{fig:reconciliation_rRMSE_ts}
\end{figure}

The average relative change of RMSE over all the time series, as well as for the whole aggregate, as a function of the step ahead are shown in Fig.~\ref{fig:reconciliation_rRMSE_ts}. Once again, it is clear that the reconciliation affects the aggregated time series positively, while it has a lower impact on the bottom one. 

\section{Conclusions}
We have discussed the performance of different forecasters and reconciliation methods in forecasting the active power demand at different levels of aggregation in an LV distribution network. We considered 24 time series (with an average power consumption of 81 kW) from a real distribution system, which were aggregated synthetically to form 7 series with two levels of aggregation. The forecaster adopted to the test the reconciliation performance was LightGBM, that, as presented in the paper, was selected as the best performing forecaster among Holt-Winters, ARMAX, KNN, and LightGBM. The reconciliation strategies that were tested are minT and Bayesian, each coupled with two methods for the estimation of the covariance matrix, i.e., graphical Lasso and the shrunk method, for a total of 4 models. Results showed that the best performing model using both upper and lower level measurements is minT with the graphical Lasso covariance estimation method. Upper-level forecasts showed the largest margin of improvements, with a reduction of the RMSE of up to 10\%. Reconciliation marginally improved the forecasting performance for the lower time series, that improved by less than 1\% on average. However, we see the possibility of using innovative reconciliation techniques, as a promising direction for future works. In particular, reconciliation can be made conditional on the forecasting errors of the base forecasters in the previous steps.

\appendices
\section{Dataset} \label{app:dataset}
The dataset consists of measurements coming from 62 IEC 61000-4-30 Class A power quality meters manufactured by DEPsys (Puidoux, Switzerland) installed in secondary substations and LV cabinets of the distribution grid of the city of Rolle (Basel, Switzerland). The dataset has been enriched with numerical weather predictions from commercial provider Meteoblue (Switzerland), updated every 12 hours. The series available in the data set along with their sampling time are reported in Table \ref{table1}. The power measurements include mean active and reactive power, voltage magnitude and maximum total harmonic distortion (THD) for each phase, voltage frequency $\omega$ and the average power over the three phases, $P_{mean}$. The latter one has been used as target variable in this paper. The meteorological forecasts include the temperature, global horizontal and normal irradiance (GHI and GNI, respectively), the relative humidity (RH) pressure and wind speed and direction ($W_s$ and $W_{dir}$, respectively).


\begin{table}[!ht]
    \renewcommand{\arraystretch}{1.3}
    \centering
    \caption{Variables available in the grid and NWP datasets}
    \label{table1}
    \begin{tabular}{cc|c|c|l}
        \cline{3-4}
        & & \multicolumn{2}{ c| }{properties} \\ \cline{3-4}
        & & variables & sampling time  \\ \cline{1-4}
        \multicolumn{1}{ |c  }{\multirow{2}{*}{\STAB{\rotatebox[origin=c]{90}{Datasets $\quad$}} }} &
        \multicolumn{1}{ |c| }{\STAB{\rotatebox[origin=c]{90}{ Grid }}} 
        & $\begin{matrix}
            P,Q,  \vert V\vert, THD \ \mathrm{(each \ phase)} \\ 
            \omega, P_{mean} 
        \end{matrix}$
        & 10 min \\ \cline{2-4}
        \multicolumn{1}{ |c  }{}                        &
        \multicolumn{1}{ |c| }{\STAB{\rotatebox[origin=c]{90}{ NWP }}} 
        & $\begin{matrix}
        T,GHI,GNI &\\
        RH,p,W_{s},W_{dir}&
        \end{matrix}$
        & 1 h, 12 h updates   \\ \cline{1-4}
    \end{tabular}
\end{table}

\begin{figure}[!ht]
	\centering
	\includegraphics[width=3.5in]{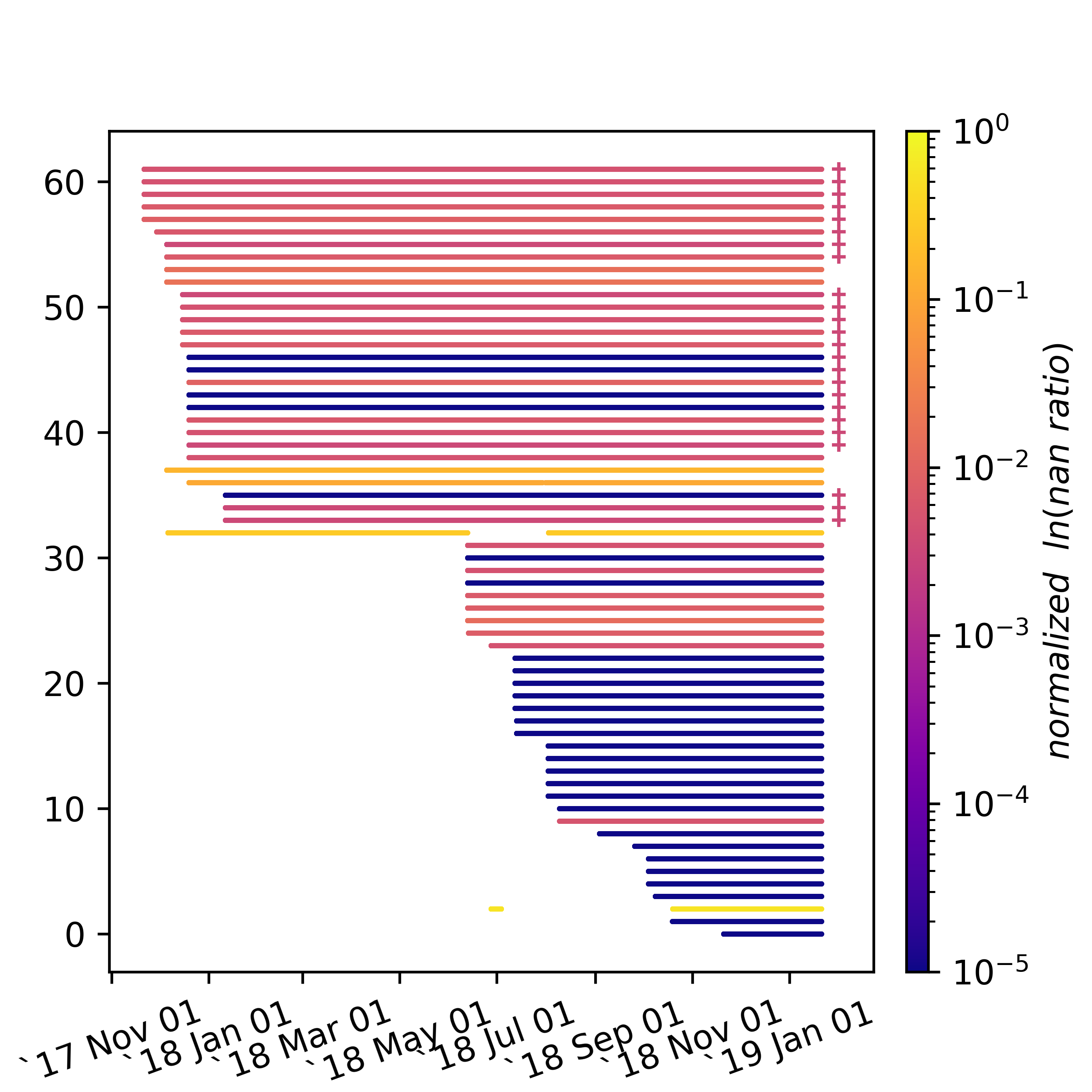}
	\caption{Time periods where the original series are present. Series are ranked by the number of available data points in descending order. Color: logarithm of the ratio of missing values from the timestamp of the first available measurement, normalized with the highest value among all the series. The red cross indicates the selected time series.}
	\label{fig:observations}
\end{figure}

Since the meters have been progressively installed in the grid (see Fig.~\ref{fig:observations}), in order to obtain a complete dataset, the meters with less than one year data have been discarded, as well as the meters presenting more than six consecutive missing values (corresponding to 1 hour). 
The remaining missing values were completed with PCHIP interpolation, which uses non-overshooting splines \cite{Fritsch2005}. Two of the selected meters presented a single sudden change of sign in the power measurements, which has been manually corrected. The final data set spans one entire year, with measurements from 13 January 2018 to 19 January 2019. Additionally to the described data, for the model for which it was possible, we also considered holidays as categorical features. These dates are available at \cite{public_dataset} as well.

{\footnotesize

}

\end{document}